\documentclass[10pt,twocolumn,letterpaper]{article}

\usepackage{cvpr}              % To produce the CAMERA-READY version

\newcommand{\tali}[2][]{%
    \ifthenelse{ \equal{#1}{} }
        {\textcolor{magenta}{(TD) #2}}
        {\textcolor{magenta}{(TD) \sout{#1}\xspace{}#2}}
}
\newcommand{\omer}[2][]{%
    \ifthenelse{ \equal{#1}{} }
        {\textcolor{orange}{(OB) #2}}
        {\textcolor{orange}{(OB) \sout{#1}\xspace{}#2}}
}

\newcommand{\yk}[2][]{%
    \ifthenelse{ \equal{#1}{} }
        {\textcolor{yonis_color}{(Yoni) #2}}
        {\textcolor{yonis_color}{(Yoni) \sout{#1}\xspace{}#2}}
}

\newcommand{\rafail}[2][]{%
    \ifthenelse{ \equal{#1}{} }
        {\textcolor{blue}{(Rafail) #2}}
        {\textcolor{blue}{(Rafail) \sout{#1}\xspace{}#2}}
}

\newcommand{\danah}[2][]{%
    \ifthenelse{ \equal{#1}{} }
        {\textcolor{blue}{(Danah) #2}}
        {\textcolor{blue}{(Danah) \sout{#1}\xspace{}#2}}
}

\usepackage[dvipsnames]{xcolor}

\newcommand{\bs}[1]{{\boldsymbol{#1}}}

\def\smmo{\texttt{SMM}}
\def\fsmm{\texttt{SMM}[\boldsymbol{f}]}

\newcommand{\myparagraph}[1]{\vspace{0.15cm}\noindent{\bf #1}\hspace{0.05cm}}
\newcommand{\afterfigure}{\vspace{-0.85em}}
\newcommand{\mytextapprox}{\raisebox{0.5ex}{\texttildelow}}

\definecolor{oursblue}{rgb}{0.21,0.49,0.74}
\usepackage[pagebackref,breaklinks,colorlinks,citecolor=oursblue]{hyperref}

\usepackage{times}
\usepackage{epsfig}
\usepackage{graphicx}
\usepackage{amsmath}
\usepackage{amssymb}
\usepackage{xcolor}
\usepackage{tabularx}
\usepackage{bbm}
\usepackage{capt-of}
\usepackage{adjustbox}
\usepackage{wrapfig}
\usepackage{algorithmic}
\usepackage{algorithm}

\definecolor{myred}{RGB}{190,20,20}

\definecolor{yonis_color}{RGB}{0,180,180}
\usepackage[normalem]{ulem}
\usepackage{ifthen}

\usepackage{xspace}
\usepackage{enumitem}

\title{Space-Time Diffusion Features for Zero-Shot Text-Driven Motion Transfer}
\vspace{-0.15cm}
\author{Danah Yatim$^{1*}$\qquad Rafail Fridman$^{1*}$\qquad Omer Bar-Tal$^{1}$ \qquad Yoni Kasten$^{2}$ \qquad Tali Dekel$^{1}$ \\
{Weizmann Institute of Science$^{1}$ \qquad NVIDIA Research$^{2}$} \\
{\small *Indicates equal contribution.} \\
}
\begin{document}
\twocolumn[{
\renewcommand\twocolumn[1][]{#1}
\maketitle
\centering
\vspace*{-0.7cm}
\includegraphics[width=1\textwidth,height=0.3\textheight]{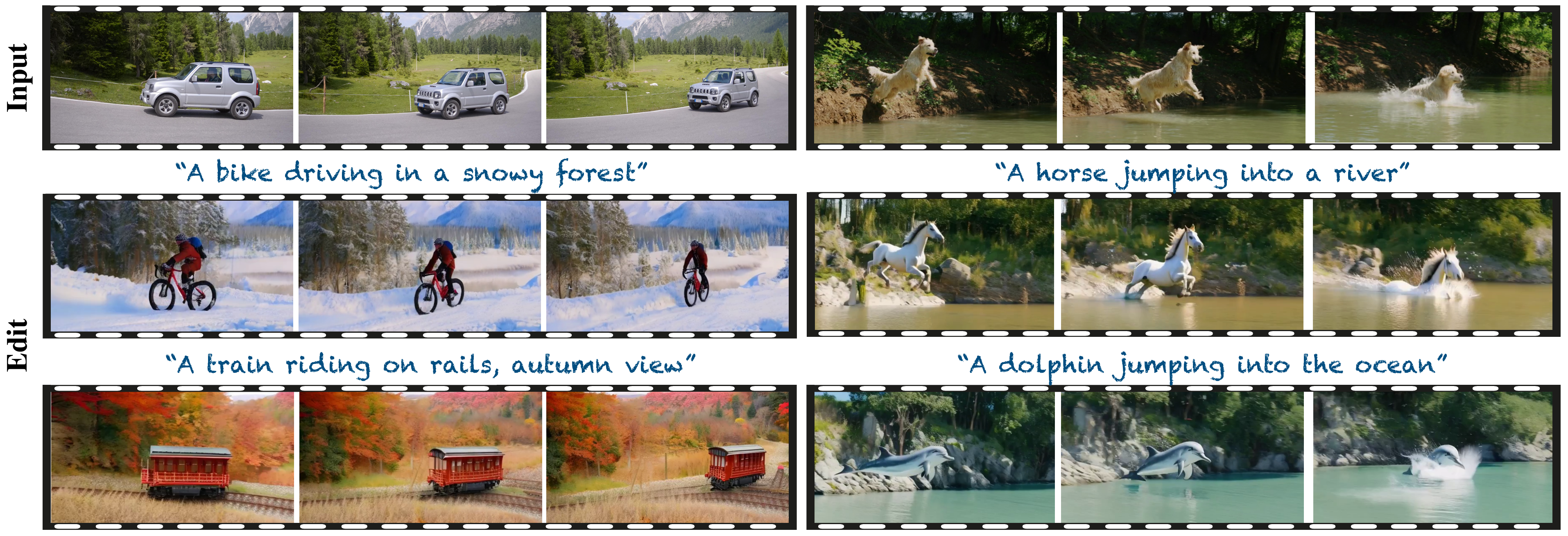} \vspace{-0.5cm}
\captionof{figure}{Given an input video and a text prompt describing the target objects and scene, our method generates a new video in which the overall motion and scene layout of the input video are preserved, while allowing for notable structural and appearance changes. }\label{fig:teaser}
 \vspace*{0.35cm}
}]

%%%%%%%%% ABSTRACT
\maketitle
\begin{abstract}
\vspace{-0.35cm}
We present a new method for text-driven motion transfer -- synthesizing a video that complies with an input text prompt describing the target objects and scene while maintaining an input video's motion and scene layout. Prior methods are confined to transferring motion across two subjects within the same or closely related object categories and are applicable for limited domains (e.g., humans). 
In this work, we consider a significantly more challenging setting in which the target and source objects differ drastically in shape and fine-grained motion characteristics (e.g., translating a jumping dog into a dolphin).  To this end, we leverage a pre-trained and fixed text-to-video diffusion model, which provides us with generative and motion priors. The pillar of our method is a new space-time feature loss derived directly from the model. This loss guides the generation process to preserve the overall motion of the input video while complying with the target object in terms of shape and fine-grained motion traits. 
\footnote{Code will be made publicly available.} 
Project page: \url{https://diffusion-motion-transfer.github.io/}
\vspace*{-2.45mm}
\end{abstract}

\section{Introduction}
\label{sec:intro}
\vspace{-0.25cm}
Imagine transferring the car's motion shown in the video in Fig.~\ref{fig:teaser} to a different object, such as a bicycle or a train. This task poses a crucial challenge --  retaining key motion characteristics of the input video, such as the car's direction of motion, speed, and pose, while substantially altering the dynamic object's structure to convey the target's unique visual attributes. Conceptually, solving this task requires prior knowledge about objects' appearance, pose, shape, and motion under deformations and different viewpoints, as well as an understanding of their interaction with the scene (e.g., turning the car into a bicycle requires revealing unseen background content and synthesizing plausible scene effects, such as shadows).

  Previous methods have been primarily focused on transferring motion across two similarly-looking videos depicting two subjects within the same object category. These methods are typically confined to object categories for which strong geometric priors exist, e.g., humans for which parametric models for robust shape and pose are available (e.g., \cite{cao2017realtime,chan2019everybody}). Other works attempt to learn such explicit mid-level representation in a self-supervised manner from the input videos (e.g., \cite{mokady2021jokr,Tao_2022_CVPR})-- an extremely challenging task by itself. In this work, we take the task of motion transfer to the realm of text-driven video editing, where the target object and scene are specified via a text prompt. We aim at addressing a significantly more general setting, which involves transferring motion across different object categories under significant variations in shape and deformations across time (Fig.~\ref{fig:teaser})

Our approach diverges from traditional motion transfer works by avoiding explicit mid-level modeling of pose and shape. Instead, we harness the \emph{generative motion priors} learned from broad video data by a pre-trained and fixed \emph{text-to-video model}. Specifically, we delve into the intermediate \emph{space-time feature representation} learned by the video model and introduce a new loss that guides the generation process of the target video towards the preservation of the original video's overall scene layout and motion. Importantly, our method allows flexibility and deviations from the exact structure and shape of the source objects. This contrasts with prior works in text-driven image and video editing that manipulate \emph{spatial features} of a pre-trained \emph{text-to-image} model. These methods inherently lack the ability to perform consistent structural edits across frames since they rely solely on spatial image priors.  To the best of our knowledge, our work is the first to empirically analyze and harness \emph{space-time} features of a text-to-video model. 
  
  Our lightweight framework works in a zero-shot fashion, requiring no training data or fine-tuning. We demonstrate results on many videos and edits, encompassing various objects and scenes.  We further suggest a new metric to measure motion similarity under shape deviation and quantitatively evaluate our method w.r.t. existing text-to-video methods, demonstrating state-of-the-art performance in terms of motion preservation and edit fidelity.   

  To summarize, our key contributions include:
  \begin{itemize}
      \item An effective zero-shot framework that harnesses the generative motion prior of a pre-trained text-to-video model for the task of motion transfer.
      \item New insights about the \emph{space-time} intermediate features learned by a pre-trained text-to-video diffusion model.
      \item A new metric for evaluating motion fidelity under structural deviations between two videos. 
      \item State-of-the-art results compared to competing methods, achieving a significantly better balance between motion preservation and fidelity to the target prompt.
  \end{itemize}

\section{Related works}
\label{sec:related}

\myparagraph{Motion transfer.}
A  related task to ours is motion transfer from a source to a target subject, where the subjects are of the same or closely related object categories (e.g., \cite{chan2019everybody,balakrishnan2018synthesizing,wei2021c2f,siarohin2019animating,siarohin2019first,wiles2018x2face,kim2018deep}). These methods take as input a driving video depicting the source motion, and an image or a video, depicting the target subject. A prevalent approach among these methods is to explicitly model the pose of the dynamic object via a parametric model (e.g.~OpenPose~\cite{cao2017realtime}). Thus, these methods are largely restricted to domains for which robust parametric models exist (e.g., humans or faces) or to transferring motion across videos depicting similar motion and closely related object categories. In contrast, we are aimed at \emph{text-driven motion transfer} across \emph{distinct object categories}. That is, the target object and scene are specified through a text prompt, where the source and target objects can differ significantly in shape, appearance, and their natural fine-grained motion traits.

\begin{figure*}[t!]
    \centering
    \includegraphics[width=\textwidth]{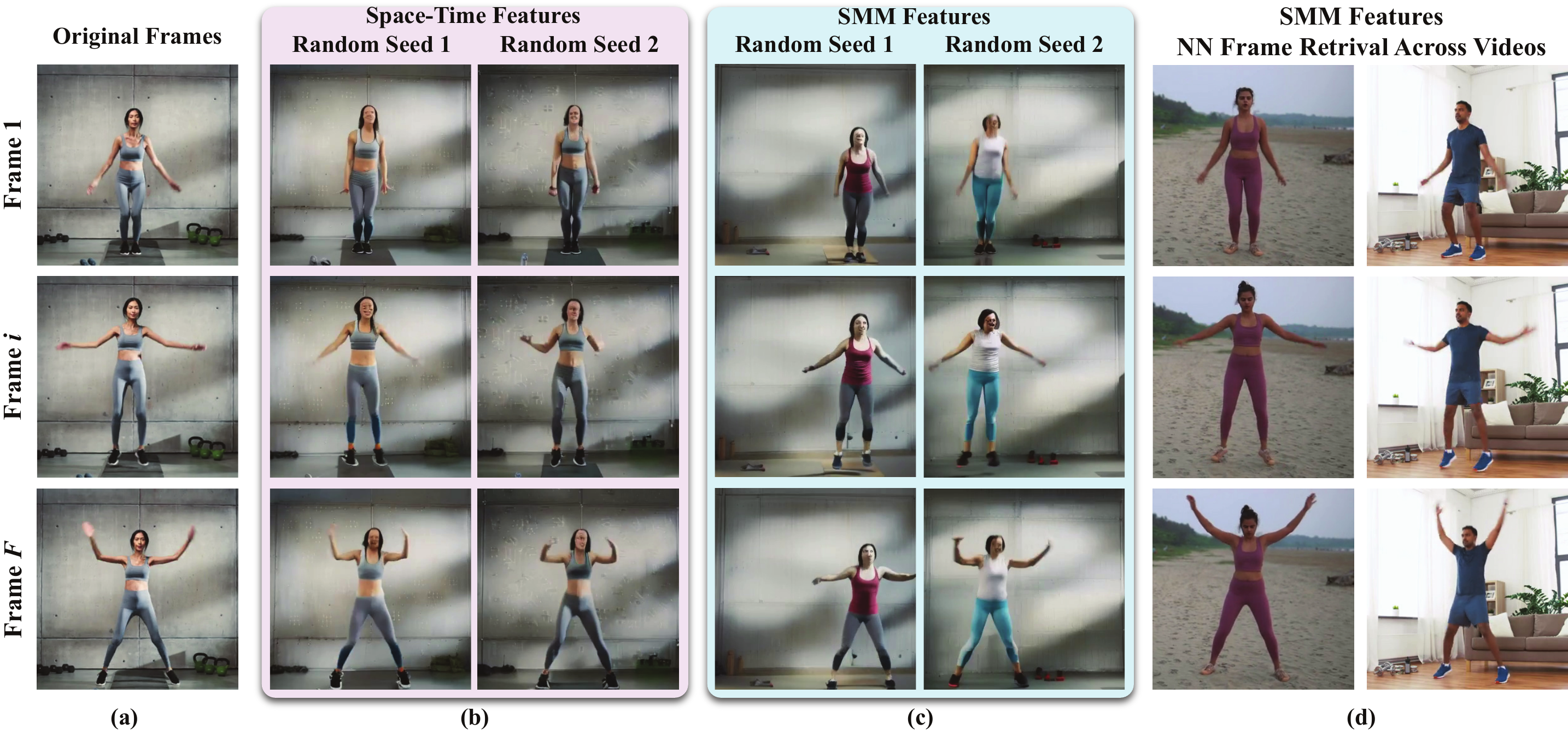} \vspace{-0.7cm}
    \caption{{\bf Diffusion feature inversion via guided feature reconstruction.} We extract space-time features $\bs{f}$ from an input video (a) and steer the generation process of a random sample to produce the same feature $\bs{f}$, using feature reconstruction as guidance (b); the synthesized videos closely resemble the original video content in terms of appearance, shape, and pose. Replacing the full space-time features with their spatial marginal mean feature $\fsmm$ allows for more flexibility (c); the SMM feature inversion results capture the original object pose, general position, and scene layout yet are not restricted to the original content at the pixel-level. This is also demonstrated in the nearest neighbor frames retrieved from other videos depicting similar actions, according to similarly in $\fsmm$ features (c).}
    \label{fig:analysis}\afterfigure
\end{figure*}

\myparagraph{Text-to-video models.}
Early works on text-to-video generation utilized VAE or conditional GAN frameworks \cite{mittal2017sync,pan2017create,li2018video} trained on small-scale datasets of simple domains (e.g., moving digits). 
Recently, there have been substantial efforts in training large-scale text-to-video models on vast datasets with autoregressive Transformers (e.g., \cite{villegas2022phenaki,wu2022nuwa,hong2022cogvideo}) or Diffusion Models (e.g., \cite{blattmann2023videoldm,ho2022imagen_video}). A prominent recent trend extends pre-trained text-to-image (T2I) diffusion models to text-to-video (T2V) generation by adding temporal layers on top of an image architecture \cite{make_a_video,blattmann2023align,ge2023preserve,po2023state}. Make-A-Video \cite{make_a_video} add temporal Convs and attention layers to a pre-trained T2I pixel-space diffusion model. Other works extend T2I diffusion models that operate on a latent space (e.g., StableDiffusion \cite{rombach2022high}) to the temporal domain \cite{blattmann2023align,ge2023preserve}. 

Several works \cite{gen1,chen2023controlavideo,wang2023videocomposer,molad2023dreamix} train or fine-tune diffusion models for video-to-video translation tasks. Gen-1 \cite{gen1} design a video architecture that is conditioned on structure/appearance representations, allowing text-driven appearance manipulation of a reference video. Control-A-Video \cite{chen2023controlavideo} extends a conditional T2I to the temporal domain, allowing the generation of videos that preserve the per-frame layout of a reference video. Nevertheless, since these methods are conditioned on generic mid-level representations of the reference video, they do not allow the preservation of the motion of the reference video \emph{while significantly deviating from the per-frame structural layout}.
In this work, we utilize a pre-trained publicly available T2V diffusion model \cite{wang2023modelscope,zeroscope} and show for the first time how it can be leveraged for motion transfer in a zero-shot manner. 

\myparagraph{Image \& video editing via feature manipulation.}
There has been unprecedented progress in text-to-image generation using diffusion models \cite{croitoru2022diffusion,beatgan,ddpm,nichol2021improved}. Following this success, a surge of works empirically analyzed the internal feature representation of prominent T2I diffusion models, e.g., StableDiffusion \cite{rombach2022high}, and showed how to perform various editing tasks using simple operations in the T2I feature space.\cite{chefer2023attend,hong2022improving,ma2023directed,pnpDiffusion2023,p2p,patashnik2023localizing,cao2023masactrl}. Prompt-to-Prompt \cite{p2p} analyzed the cross-attention maps and showed how to manipulate them for controlling the spatial composition in generated images. Plug-and-Play Diffusion (PnP) \cite{pnpDiffusion2023} showed that the spatial features capture semantic information at high spatial granularity and utilized them for image-to-image translation. 

A prominent line of works adopt a pre-trained T2I diffusion model for \emph{video} editing \cite{tokenflow2023,wu2022tuneavideo,qi2023fatezero,Ceylan2023Pix2VideoVE,text2video-zero}. For example, Tune-A-Video \cite{wu2022tuneavideo} fine-tunes a T2I model on a single input video and uses the fine-tuned model for stylizing the video or replacing object categories. TokenFlow \cite{tokenflow2023} performs consistent video editing in a zero-shot manner by enforcing consistency on the diffusion features across frames. Nevertheless, these methods do not have access to a T2V generative motion prior and are not designed for edits that require significant structural deviation from the original video. Unlike these works, we leverage the motion prior of a large-scale T2V model, which allows us to support edits that require shape deviation (e.g., car to bike in Fig. \ref{fig:teaser}), utilizing a novel loss function that we derive from the model's feature space. To the best of our knowledge, our work is the first to investigate the internal feature representation of a T2V model and leverage it for an editing task.

\myparagraph{Consistent video editing.} Several recent methods have leveraged layered video representations for consistent text-driven video editing \cite{bar2022text2live,kasten2021layered,jafarian2023normal,lee2023shape,ye2022deformable,chai2023stablevideo,loeschcke2022text}.  For example, Text2LIVE \cite{bar2022text2live} proposed to use a pre-trained Layered Neural Atlases (NLA) \cite{kasten2021layered} representation, which is edited using losses defined in CLIP text-image space \cite{clip}. However, since the structure and motion of the edited video are determined via the pre-trained NLA, such methods are restricted to only appearance changes. Recently, \cite{lee2023shape} generalized this approach to allow local structural changes in the edited videos. Nevertheless,  NLA takes hours to train, and cannot faithfully represent complex videos due to the strong regularization of objects' motion.  In addition, all these methods use a 2D generative prior, thus cannot support large structure deviation and adaption of motion.

\begin{figure*}[t!]
    \centering
    \includegraphics[width=1\textwidth]{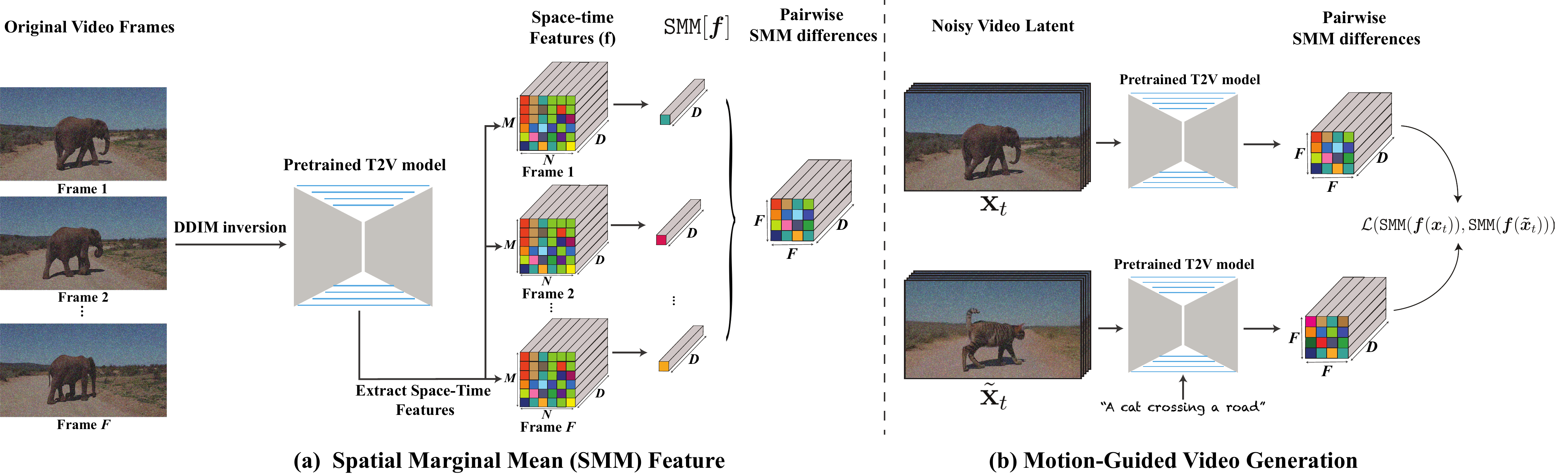}
    \caption{{\bf Pipeline.} (a)  Given an input video, we apply DDIM inversion and extract space-time features $\bs{f}\in \mathbb{R}^{F\times M \times N \times D}$ from intermediate layer activations. We obtain our Spatial Marginal Mean (SMM) feature $\fsmm \in \mathbb{R}^{F \times D}$ by computing the mean over the spatial dimensions, and compute the pairwise differences between each pair of SMM features. (b) For editing, we guide the generation at each denoising step with our Pairwise SMM differences objective (b). See Sec.~\ref{sec:method} for more details.} \afterfigure
    \label{fig:pipeline}
\end{figure*}

\section{Preliminary}
\label{sec:prelim}
\myparagraph{Diffusion models.}
\label{sec:prelim_dm}
Diffusion models \cite{ddpm,beatgan,sohl2015deep,po2023state} are generative models that aim to approximate a data distribution $q$ by mapping an input noise
$\bs{x}_T\sim \mathcal{N}(0,I)$ to a clean sample $\bs{x}_0\sim q$, through an iterative denoising process. 
The DDIM sampler allows to denoise an initial noise $\bs{x}_T$ in a deterministic manner \cite{song2020_ddim}. By applying DDIM inversion, a clean sample $\bs{x}_0$ can be mapped back to the sequence of noisy samples $\{\bs{x}_i\}_{t=1}^T$ used to generate it. 

In latent text-to-image (T2I) diffusion models, e.g., StableDiffusion \cite{rombach2022high},  a pre-trained encoder compresses an RGB image  $\bs{I} \in \mathbb{R}^{H \times W \times 3}$ to a latent  $\bs{x} \in \mathbb{R}^{H' \times W' \times 4}$, which can be decoded back to the high-resolution space. These T2I models comprise a \mbox{UNet} architecture \cite{ronneberger2015unet}, which consists of Convolutions and Attention modules. 

Latent video diffusion models (e.g., \cite{blattmann2023align,wang2023modelscope}) extend latent T2I models to text-to-video (T2V) by inflating the 2D architecture to the temporal domain, i.e., adding temporal convolutions and temporal attention layers, and fine-tuning on video datasets. In this case, the T2V model generates a latent video $\bs{x} \in \mathbb{R}^{F \times H' \times W' \times 4}$, which is then decoded to the output RGB video $\mathcal{V}\in \mathbb{R}^{F \times H \times W \times 3}$. 
In this work, we utilize the publicly available ZeroScope T2V \cite{zeroscope} model, which inflates StableDiffusion \cite{rombach2022high}. 

\section{Method}
\label{sec:method}

Given an input video $\mathcal{V}$ and a target text prompt $P$, our goal is to generate a new video $\mathcal{J}$ that preserves the overall motion and semantic layout of $\mathcal{V}$, while complying with $P$. We utilize  ZeroScope --  a pre-trained latent T2V model \cite{zeroscope,wang2023modelscope}. 

The key component of our method, illustrated in Fig.~\ref{fig:pipeline}, is a novel objective function that is used as guidance during the generation process of $\mathcal{J}$. We conceive this objective based on our empirical analysis that reveals new insights about space-time diffusion features extracted from the model. Specifically, we show that the first-order statistics of the features in the spatial dimensions, i.e., the spatial marginal mean of the features, can serve as a powerful per-frame global descriptor that (i) retains spatial information such as objects' position, pose, and the semantic layout of the scene, and (ii) robust to pixel-level variations in both appearance and shape.

\subsection{Space-time analysis of diffusion features}
\label{sec:anaylsis}
We focus our analysis on features extracted from the intermediate layer activations of the video model. Recall that the video model is initialized from a text-to-image model, for which the semantic DIFT features \cite{tang2023emergent}, were shown to encode localized \emph{semantic} information shared across objects from different domains \cite{pnpDiffusion2023,tang2023emergent}. Here, we examine the \emph{space-time} properties of the corresponding features in the \emph{video model}. See Supplementary Material (SM) for details about the video model architecture and feature selection.

Given the input video $\mathcal{V}$, we apply DDIM inversion with an empty prompt \cite{song2020_ddim}, and obtain a sequence of latents $[\bs{x}_1,...,\bs{x}_T]$, where $\bs{x}_t$ is the video latent at generation step  $t$. We input the latent $\bs{x}_t$ to the network and extract the space-time features $\bs{f}(\bs{x}_t) \in \mathbb{R}^{F \times M \times N \times D}$, where $F, M, N$ are the number of frames, height and width of the $D$ dimensional feature activation, respectively.

\myparagraph{Diffusion feature inversion.} To gain a better understanding of what the features $\{\bs{f}(\bs{x}_t)\}_{t=1}^T$  encode, we adopt the concept of ``feature inversion'' \cite{simonyan2013deep,mahendran2015understanding,tumanyan2022splicing}. Our goal is to optimize for a video $\mathcal{V}^*$, randomly initialized, that would produce these features when fed into the network. Specifically, this is achieved using feature reconstruction guidance \cite{mou2023dragondiffusion,epstein2023selfguidance} during the sampling process of $\mathcal{V}^*$. Formally,
\vspace{-0.2cm}
\begin{equation}
    \begin{array}{l}
    \bs{\hat{x}}_T \sim \mathcal{N}(0,\mathcal{I}) \\
    \bs{\hat{x}}_{t-1} \!=\! \Phi(\bs{x}^*_t, P_s),\text{where}~~ \nonumber 
    \bs{x}^*_t \!=\! \operatorname*{argmin}_{\mathbf{\hat{\bs{x}}}} \|\mathbf{\bs{f}(\bs{x}_t)}-\mathbf{\bs{f}(\hat{\bs{x}}_t)}\|^2
    \end{array}
    \label{eq:inversion}
\end{equation}
Here, $\Phi$ is the diffusion model, and $P_s$ is a general text prompt describing the input video (e.g., ``a car''). We minimize the feature reconstruction objective using gradient descent at each generation step. See SM for more details.  

Figure~\ref{fig:analysis} shows our inversion results for the space-time features extracted from an input video; we repeat the inversion process several times, each with different random initialization (i.e., different seeds). We observe that inverted videos nearly reconstruct the original frames  (Fig.~\ref{fig:analysis}(b)). 

Ultimately, we opt to find a feature descriptor that retains such information about objects' pose and the semantic layout of the scene yet is robust to variations in appearance and shape. To reduce the dependency on the pixel-level information,  we introduce a new feature descriptor, dubbed \emph{Spatial Marginal Mean (SMM)}, obtained by reducing the spatial dimensions. Formally, 
\vspace{-0.3cm}
\begin{equation} \label{eq:smm} 
    \texttt{SMM}[\bs{f}(\bs{x}_t)] = \frac{1}{M\cdot N} \sum_{i=1}^M \sum_{j=1}^N \bs{f}(\bs{x}_t)_{i,j} \afterfigure
\end{equation}
Where $\bs{f}(\bs{x}_t)_{i,j} \in \mathbb{R}^D$ is the entry at spatial location $(i,j)$ in the space-time feature volume $\bs{f}(\bs{x}_t)$. 

\begin{figure*}[ht!]
    \centering
    \includegraphics[width=\textwidth]{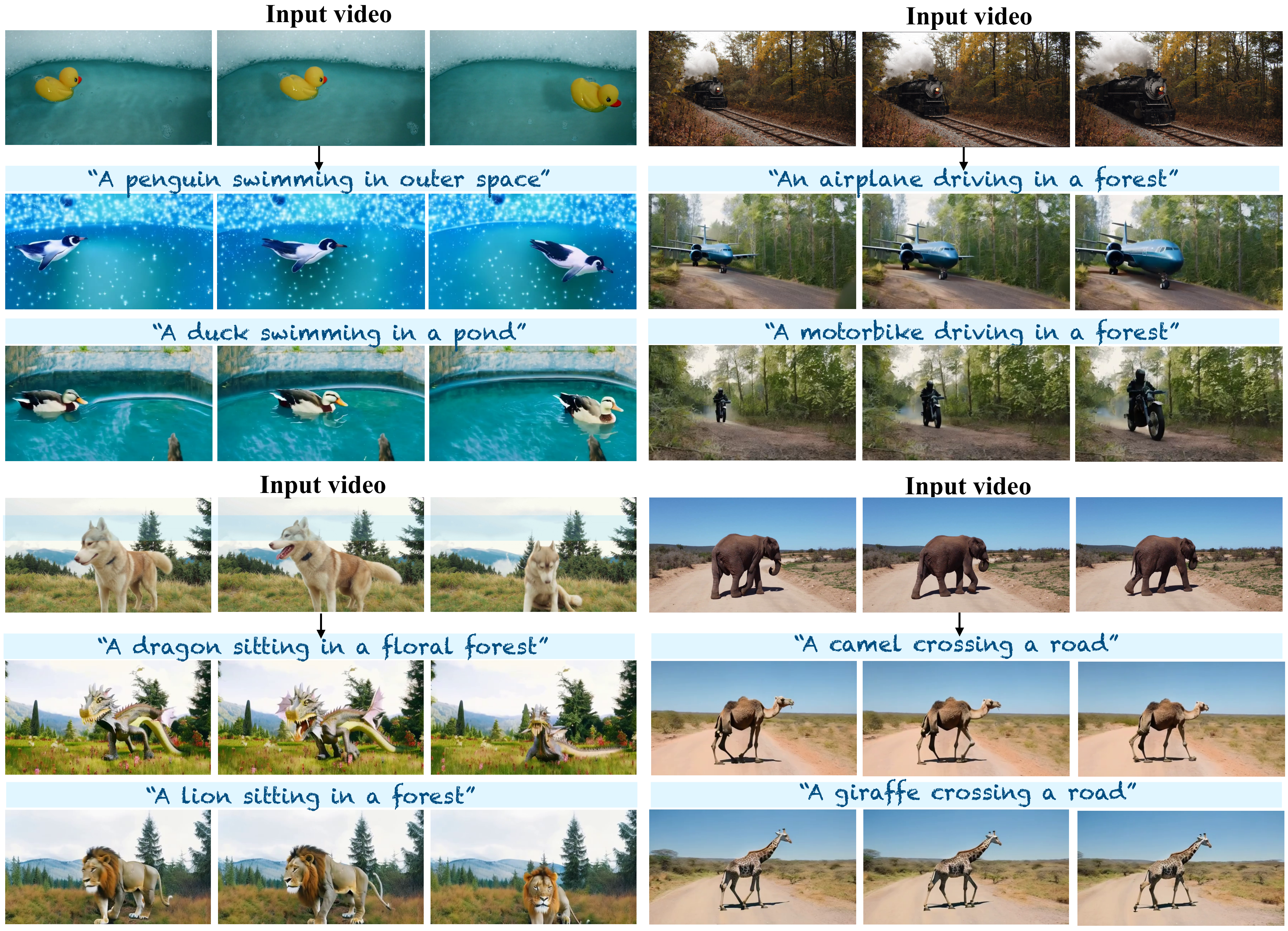} \vspace{-0.7cm}
    \caption{{\bf Sample results of our method.} See SM for full video results.}\afterfigure
    \label{fig:results}
\end{figure*}
We repeat the inversion experiment (Eq.~\ref{eq:inversion}), with  $\{\texttt{SMM}[\bs{f}(\bs{x}_t)]\}_{t=1}^T$ as the target features to reconstruct.  Figure~\ref{fig:analysis}(c) shows the inversion results for the SMM features, for different initializations. 
Remarkably, although the spatial dimensions are collapsed in the SMM features, the inverted videos convey the correct pose and position of objects, while depicting larger structural and appearance variations than using the full space-time features. 

We further demonstrate these properties by treating the spatial marginal mean associated with each frame as a global per-frame descriptor, and using it to retrieve nearest neighbour frames from other videos. As seen in Fig.~\ref{fig:analysis}(d), the retrieved nearest frames depict the same pose, under noticeable appearance and viewpoint changes.

\subsection{Motion-guided video generation}
Based on our findings, we now turn to the task of generating a new video $\mathcal{J}$ that complies with the input prompt $P$ and preserves the motion characteristics of the driving video $\mathcal{V}$. 

Our feature inversion analysis raises the question of whether the same approach can be used for editing, by simply replacing the source prompt $P_s$ with an edit prompt $P$ in  Eq.~\ref{eq:inversion}. 
 
\begin{figure*}[t!]
    \includegraphics[width=\textwidth]{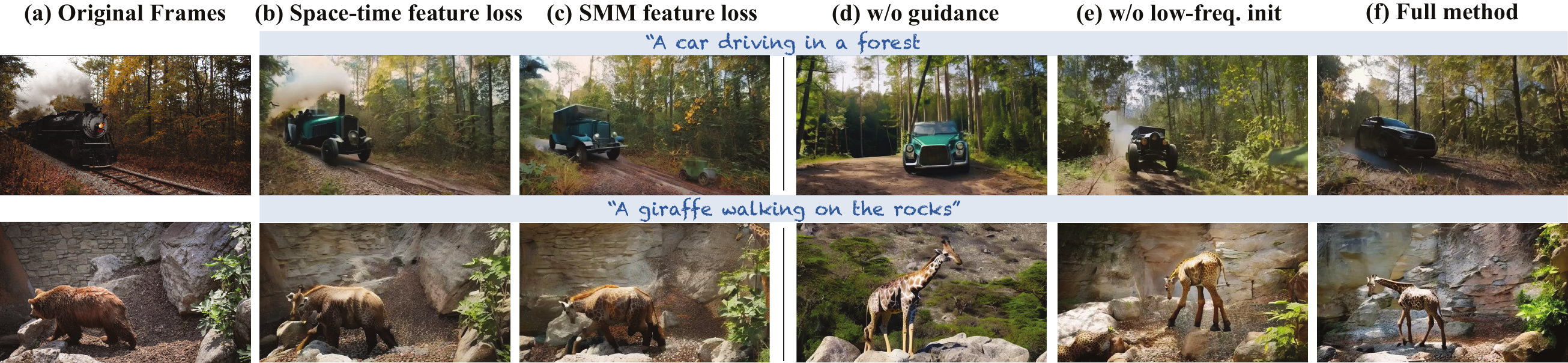} \vspace{-0.7cm}
    \caption{
    {\bf Ablation.} (b-c) We compare alternative loss functions instead of our pairwise SMM differences loss (Eq.~\ref{eq:diff_objective}); (b) using full space-time features reconstruction prevents deviations in appearance and shape; and (c) SMM feature reconstruction allows for more flexibility yet retains appearance information. (d-f) Ablation of our key components: (d) directly sampling from the initial latent w/o optimization preserves only the coarse layout. (e) Starting our optimization from randomly sampled noise (w/o low frequency filtering Eq.~\ref{eq:init}) results in lower motion fidelity compared to our full method (f).}\afterfigure 
    \label{fig:ablation}
\end{figure*}

 Figure~\ref{fig:ablation} shows these results for a couple of videos, which demonstrate two issues: (i) depending on the initialization, the optimization may converge to a local minima in which the accurate position of objects and their orientation may differ from the input, (ii) the SMM features still retain appearance information, which reduces the fidelity to the text prompt. We propose the following two components to resolve these issues.

\myparagraph{Pairwise SMM differences.} As seen in Fig.~\ref{fig:ablation}, directly optimizing for the SMM features often prevents us from deviating from the original appearance.  
To circumvent this problem, we propose an objective function that aims to preserve the pairwise differences of the SMM features, rather than their exact values. Formally,  let $\phi_i^t,\tilde{\phi}_i^t\in \mathbb{R}^d$ be the SMM features for frame $i$ and step $t$ for the driving video and the generated video, respectively. For a generation step $t$, the pairwise SMM differences $\Delta^t,\tilde{\Delta}^t \in \mathbb{R}^{F\times F\times d}$ are defined as follows:
\vspace{-0.2cm}
\begin{equation}
\Delta_{(i,j)}^t=\phi_i^t-\phi_j^t \ \ \  \ \ \ \tilde{\Delta}_{(i,j)}^t=\tilde{\phi}_i^t-\tilde{\phi}_j^t  
\end{equation}
for $i,j \in \{1,\dots,F\}$. Our loss for time step $t$ is then:
\begin{equation}
    \mathcal{L}(\smmo(\bs{f}(\bs{x}_t)), \smmo(\bs{f}(\bs{\tilde{x}}_t))) \!=\!  \sum_i \sum_j ||\Delta^t_{(i,j)}-\tilde{\Delta}^t_{(i,j)}||^2_2
\end{equation}
\label{eq:diff_objective}
Intuitively, this loss lets us preserve the relative changes in the features through time, while discarding the exact appearance information of the source video (Fig.~\ref{fig:ablation}).

\begin{algorithm}[t!]
\small
    \caption{\bf{Motion-Guided Video Generation}}
    \label{alg:method}
    \textbf{Input:}
    \begin{algorithmic}
        \STATE $\mathcal{V}, \mathcal{P}$ \hfill $\triangleright$ Input video and target text prompt
    \end{algorithmic}
    $\{ \bs{x}_t \}_{t=1}^T \gets \text{DDIM-Inv}[\mathbf{V}] \quad \forall t\in[T]$ \\
    $\epsilon_0 \sim \mathcal{N}(0,\mathcal{I})$
    \\
    $\bs{\tilde{x}}_T=LF_\xi (\bs{x}_T)+\left(\epsilon_0-LF_\xi(\epsilon_0 )\right)$ \hfill \hspace{-0.5cm} $\triangleright$  Filtered noise (Eq.~\ref{eq:init})
    \\
    \textbf{For} $t=T,\dots,1$ \textbf{do}
    \begin{algorithmic}
        \STATE $\bs{f}(\bs{x}_t), \bs{f}(\bs{\tilde{x}}_t) \gets$ Extract space-time features 
        \STATE $\smmo(\bs{f}(\bs{x}_t)), \smmo(\bs{f}(\bs{\tilde{x}}_t)) \gets$ Spatial marginal mean (Eq.~\ref{eq:smm})
        \STATE $\bs{x}^*_t = \operatorname*{argmin}_{\bs{\tilde{x}}_t} \mathcal{L}(\smmo(\bs{f}(\bs{x}_t)), \smmo(\bs{f}(\bs{\tilde{x}}_t)))$
        \STATE $\bs{\tilde{x}}_{t-1} = \Phi(\bs{x}^*_t, P)$ \hfill $\triangleright$ Apply a denoising step
    \end{algorithmic}
    \textbf{Output:} $\mathcal{J}\gets \bs{x}_0$
\end{algorithm}  

\myparagraph{Initialization.}    It is well-known that the diffusion denoising process is performed in a coarse-to-fine manner, thus, the initialization plays an important role in defining the low frequencies of the generated content \cite{meng2022sdedit,bar2023multidiffusion}. Initialization from a random point may often converge to an undesired local minimum, in which object position is not well-preserved. Note that the low-frequency information of the original video is readily available in the DDIM inverted noise $\bs{x}_T$. However, we empirically found that this initialization may often restrict edit-ability \cite{parmar2023zero} (see SM for an example).  We thus extract only the low frequencies from $\bs{{x}}_T$. Specifically, let $\bs{x}\in \mathbb{R}^{F\times M\times N   }$  be a tensor representing $F$ frames,  with a spatial resolution of $M\times N$. We denote by $LF_\xi(\bs{x})$ the operation of spatially downsampling and upsampling $\bs{x}$ by a factor of $\xi$.    Then, our initial latent $\bs{\tilde{x}}_T$ is given by: 
\vspace{-0.2cm}
\begin{equation}
    \bs{\tilde{x}}_T=LF_\xi (\bs{x}_T)+\left(\epsilon_0-LF_\xi(\epsilon_0 )\right)
\end{equation}
\label{eq:init}
where $\epsilon_0 \sim \mathcal{N}(0,I)$ is a random noise.  Intuitively, $\bs{\tilde{x}}_T$ preserves the low-frequencies of the DDIM noise where the higher frequencies are determined by $\epsilon_0$.

To summarize, starting from the filtered latent $\bs{\tilde{x}}_T$, our method deploys the following guided generation process:
\begin{equation}
    \begin{array}{l}
    \bs{x}^*_t = \operatorname*{argmin}_{\bs{\tilde{x}}_t} \mathcal{L}(\smmo(\bs{f}(\bs{x}_t)), \smmo(\bs{f}(\bs{\tilde{x}}_t))) \\ \nonumber
    \bs{\tilde{x}}_{t-1} = \Phi(\bs{x}^*_t, P) 
    \end{array}
    \label{eq:generation}
\end{equation}
Our full framework is summarized in Alg.~\ref{alg:method}.

\section{Results}
\label{sec:results}

We evaluate our method on various scenes and object categories, most of which involve camera as well as object motion. The driving videos are taken from DAVIS dataset \cite{davis} and from the Web. Our video results and implementation details are available in the Supplementary Materials (SM). 

Figures~\ref{fig:teaser},~\ref{fig:results} show sample results of our method. As seen, our method facilitates edits that involve notable changes to the shape and structure of deforming objects, while preserving camera and objects' motion.  For instance,  we preserve the 3D pose of the car when transferring its motion to a bike or a train (Fig.~\ref{fig:teaser}); and maintain the actions of non-rigidly moving objects, e.g., sitting dog or walking camel in Fig.~\ref{fig:results}.  

\myparagraph{Baselines} 
We compare our method to the following text-driven video editing methods: (i) \emph{Shape-Aware Layered Neural Atlases (SA-NLA)} \cite{lee2023shape} that utilizes a pre-trained layered video representation  \cite{kasten2021layered} and a pre-trained T2I model~\cite{rombach2022high}. (ii) \emph{TokenFlow} \cite{tokenflow2023}, a zero-shot method that works in the feature space of a pre-trained T2I model (iii) \emph{GEN-1} \cite{gen1} and (iv) \emph{Control-A-Video} \cite{chen2023controlavideo}, both are video-to-video diffusion models that condition the generation on input depth maps, (v) \emph{Tune-A-Video} \cite{wu2022tuneavideo} that inflates a T2I model and finetunes it on a single test video, and (iv) SDEdit \cite{meng2022sdedit} applied to the same T2V model as our method. 

\begin{figure}[t!]
    \centering
    \includegraphics[width=\columnwidth]{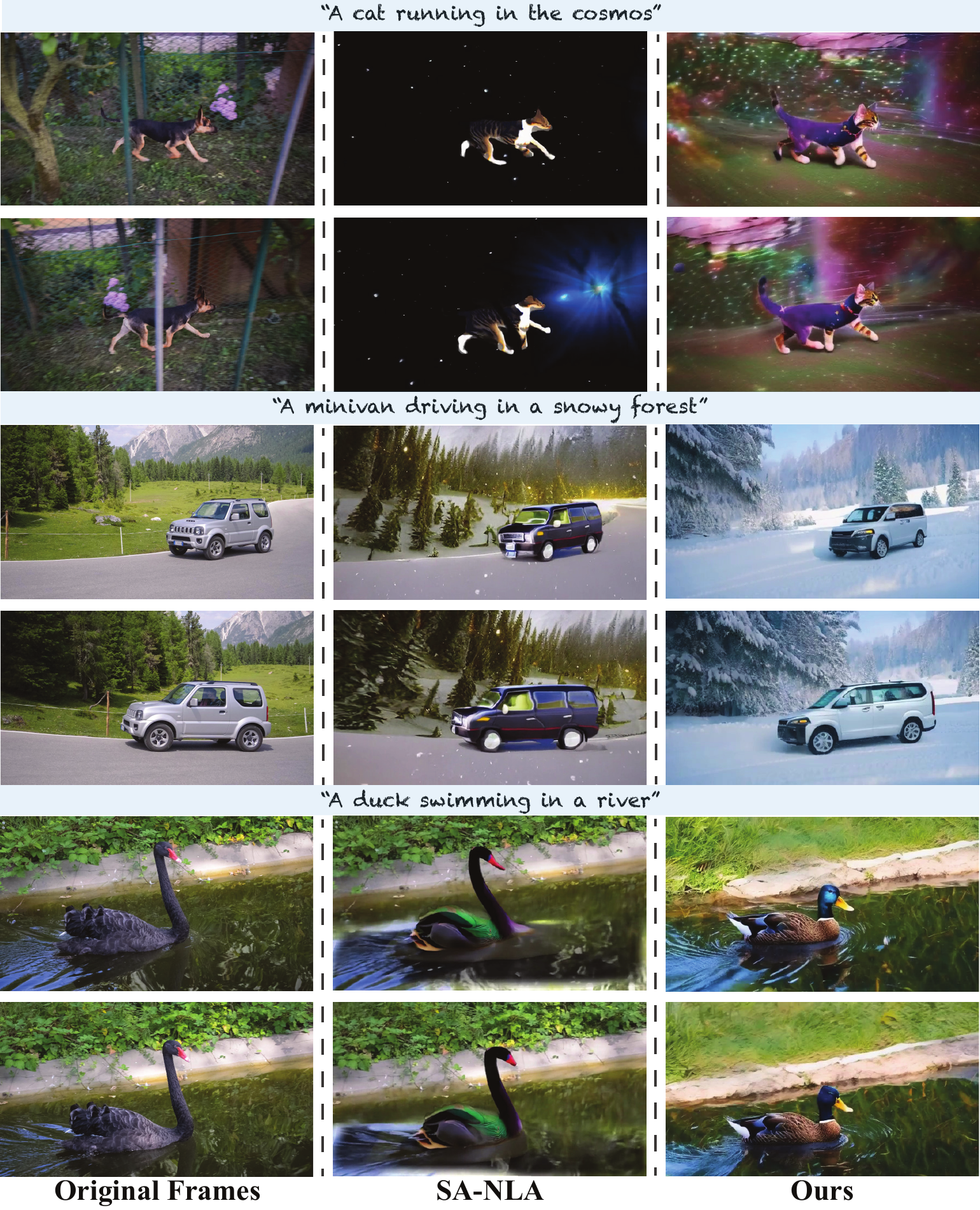} \vspace{-0.7cm}
    \caption{{\bf Comparison to SA-NLA \cite{lee2023shape}}. See SM for video results. }
    \label{fig:shapeaware}\afterfigure
\end{figure}

\begin{figure*}[t!]
    \centering
    \includegraphics[width=\textwidth]{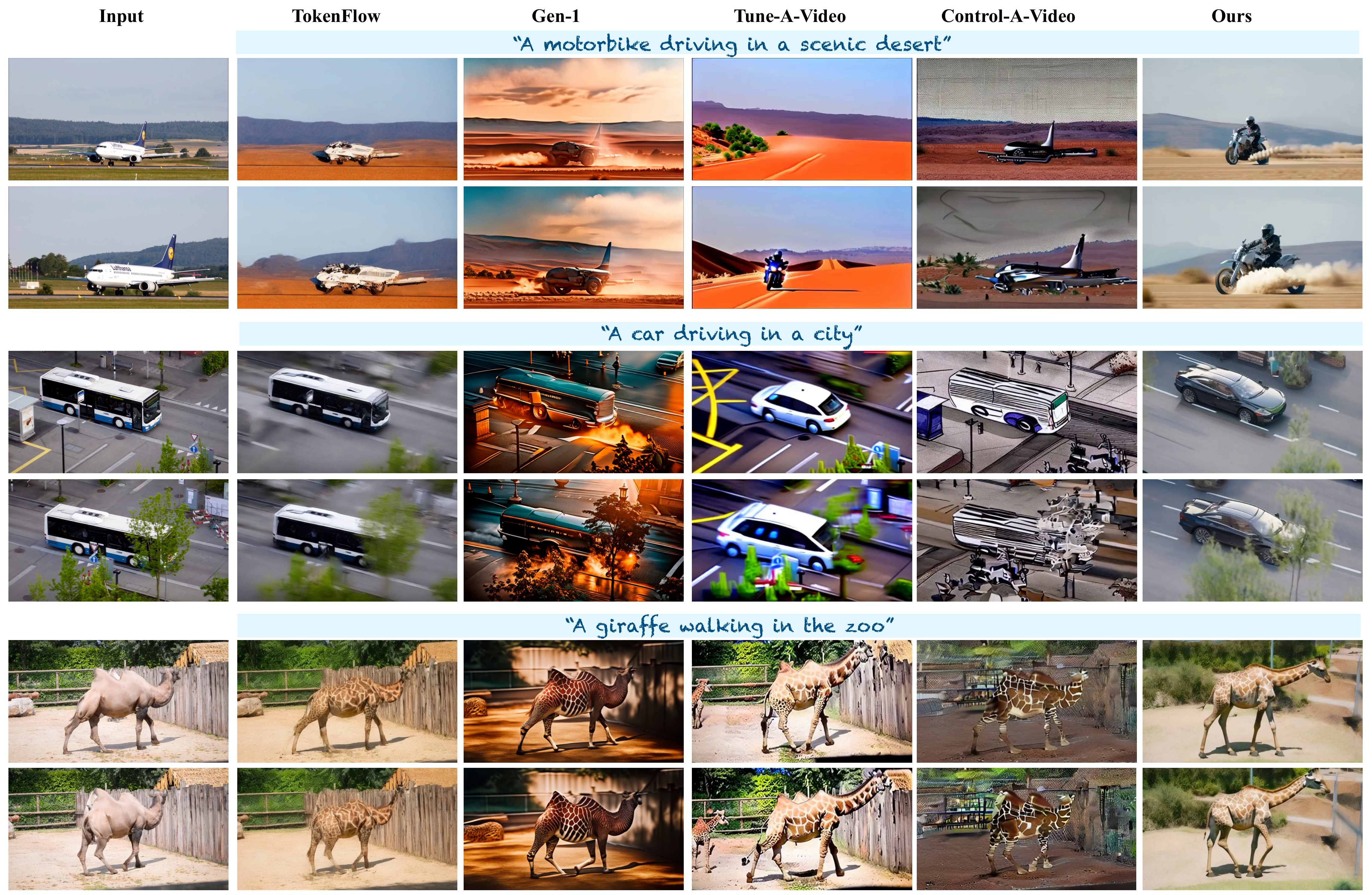} \vspace{-0.7cm}
    \caption{{\bf Comparison.} Sample results comparing our method to TokenFlow~\cite{tokenflow2023}, Gen-1~\cite{gen1}, Tune-A-Video~\cite{wu2022tuneavideo}, and Control-A-Video~\cite{chen2023controlavideo}. See SM for full video comparisons.} 
    \label{fig:comparisons}\afterfigure
\end{figure*}

\subsection{Qualitative evaluation}
Figure~\ref{fig:shapeaware} shows a qualitative comparison to SA-NLA \cite{lee2023shape}. Note that SA-NLA utilizes a layered video representation \cite{kasten2021layered}, which requires foreground/background separation masks and takes $\sim$10 hours to train. Thus, we compare to their provided videos and edit prompts qualitatively.  As seen in Fig.~\ref{fig:shapeaware},  both our method and SA-NLA exhibit high fidelity to the original motion. Nevertheless, our method allows for greater deviation in structure, (e.g., matching the structure of a duck in the swan example) and adaption of fine-grained motion traits, which are necessary for capturing the unique attributes of the target object. For example, adapting the shape and movement of a dog's tail to resemble a naturally-looking cat's tail.

Figure~\ref{fig:comparisons} provides comparisons to the additional baselines. As seen, none of these methods can both convey the original motion and adhere to the edit prompt. TokenFlow \cite{tokenflow2023} is tailored to preserve structure of the input video. Gen-1 \cite{gen1} and Control-A-Video \cite{chen2023controlavideo} struggle to deviate from the input shapes as they condition the generation on the per-frame depth maps extracted from the input video. Tune-A-Video \cite{wu2022tuneavideo} manages to fulfill the target prompt, yet objects are not aligned in pose and motion (e.g., camel to bear).  Our method significantly outperforms these baselines, by successfully matching the desired edits, which require significant structural changes and may involve synthesizing dynamic scene elements (e.g., smoke behind the motorbike). We refer the reader to the SM for full video results.

\myparagraph{Quantitative evaluation.} We numerically evaluate the key aspects of our results using the following metrics:  \vspace{0.1cm}\\
\noindent \textit{(i) Edit fidelity.} Following previous works (e.g., \cite{gen1,tokenflow2023}), we use CLIP \cite{clip} to measure the similarity between each frame and the target text and report the average score.
\vspace{0.1cm}

\noindent \textit{(ii) Motion fidelity.} We aim to assess the fidelity of our results in preserving the original motion. Given our task involves structural deviations, there is no alignment between pixels in the original and output videos. Consequentially,  traditional metrics such as comparing optical-flow fields are unsuitable for our use case. We thus introduce a new metric, based on similarity between \emph{unaligned} long-trajectories. 
We expect that even under structural changes, the two sets of trajectories would exhibit shared characteristics. 

Specifically, we use off-the-shelf tracking method \cite{karaev2023cotracker} to estimate $\mathcal{T}=\{\tau_1,\dots,\tau_n\},\tilde{\mathcal{T}}=\{\tilde{\tau}_1,\dots,\tilde{\tau}_m\}$, two sets of tracklets in the input and output videos, respectively. 
    
    Inspired by the Chamfer distance, we define our \emph{Motion-Fidelity-Score} as follows. For each tracklet $\tau_i\in \mathcal{T}$, we measure the similarity to its nearest neighbor in $\tilde{\tau}_i \in \mathcal{T}$, and vice versa. 
    \vspace{-0.1cm}
     \begin{align} 
     \frac{1}{m}\sum_{\tilde{\tau}\in \tilde{\mathcal{T}}} \underset{\tau \in \mathcal{T} }{\text{max}} \ \textbf{corr} (\tau,\tilde{\tau})+\frac{1}{n}\sum_{{\tau}\in {\mathcal{T}}} \underset{\tilde{\tau} \in \tilde{\mathcal{T}} }{\text{max}} \ \textbf{corr} (\tau,\tilde{\tau}) 
     \label{eq:motion}
     \end{align}
where the correlation between two tracklets $\textbf{corr}(\tau,\tilde{\tau})$ is computed as follows, similarly to \cite{liu2005motion}: 
    \begin{align}
    \textbf{corr}(\tau,\tilde{\tau}) = \nonumber
        \frac{1}{F}\sum_{k=1}^F\frac{{v_k^x \cdot \tilde{v}_k^x + v_k^y \cdot \tilde{v}_k^y}}{{\sqrt{(v_k^x)^2 + (v_k^y)^2} \cdot \sqrt{(\tilde{v}_k^x)^2 + (\tilde{v}_k^y)^2}}}
    \end{align}
    where $(v_k^x,v_k^y),(\tilde{v}_k^x,\tilde{v}_k^y)$ are the $k^{th}$ frame displacement of tracklets $\tau,\tilde{\tau}$ respectively.

\begin{figure}
    \centering
    \includegraphics[width=.46\textwidth]{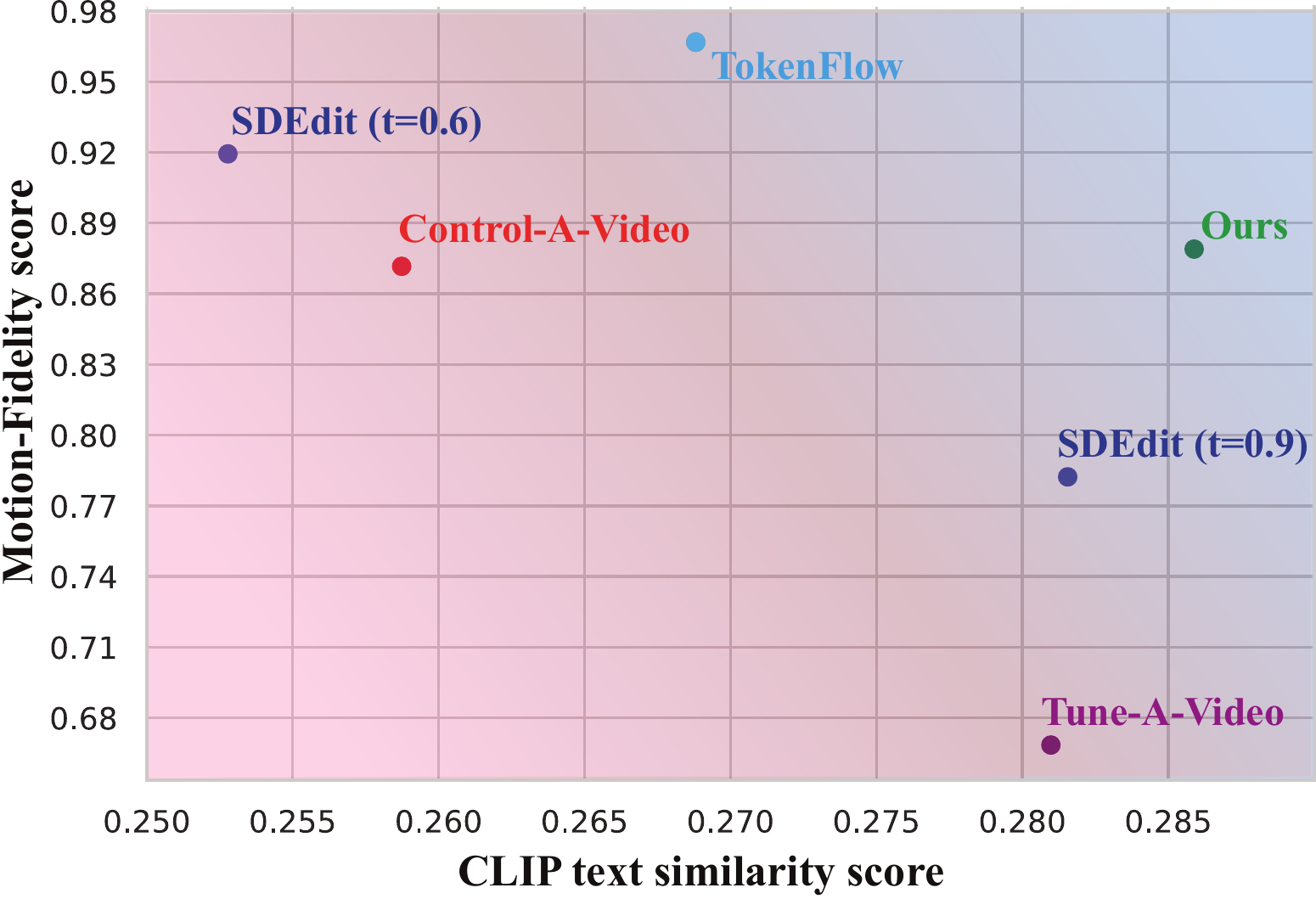}\vspace{-0.3cm}
    \caption{{\bf Quantitative evaluation.} For each baseline, we measure CLIP text similarity (higher is better) and motion fidelity (Eq.~\ref{eq:motion}; higher is better). Our method exhibits a better balance between these two metrics.}\afterfigure
    \label{fig:metrics}
\end{figure}

Figure.~\ref{fig:metrics} reports the metrics above for a set of 54 video-edit text pairs containing 21 unique videos. Our method outperforms the baselines by achieving both high fidelity to the target text prompt and the original motion. 
As expetcted, TokenFlow \cite{tokenflow2023} achieves high motion fidelity score, yet a low edit fidelity score. Control-A-Video \cite{chen2023controlavideo} exhibits a similar behaviour since it utilizes depth maps as a guidance signal to edit the video. 
Tune-A-Video \cite{wu2022tuneavideo} shows an inverse trend, i.e., satisfying the desired edit at the cost of motion fidelity. We further consider SDEdit \cite{meng2022sdedit} with different noise levels, none of which can resolve the motion-edit tradeoff.
Note that Gen1's API outputs a different number of frames, thus we could not quantitatively evaluate their performance.

\vspace{0.1cm}
\noindent (iii) \emph{User study.} We employ the Two-alternative Forced Choice (2AFC) protocol for text-driven video editing \cite{bar2022text2live,tokenflow2023,qi2023fatezero,gen1}. Participants are presented with the input video, our results and a baseline, and are asked to determine which video better aligns with the text prompt while preserving the motion of the original video.
We collected ~7000 user judgments from  150 users. As seen in Table~\ref{table:amt}, our method is consistently preferred over all baselines.

\begin{figure}[t!]
    \centering
    \includegraphics[width=.5\textwidth]{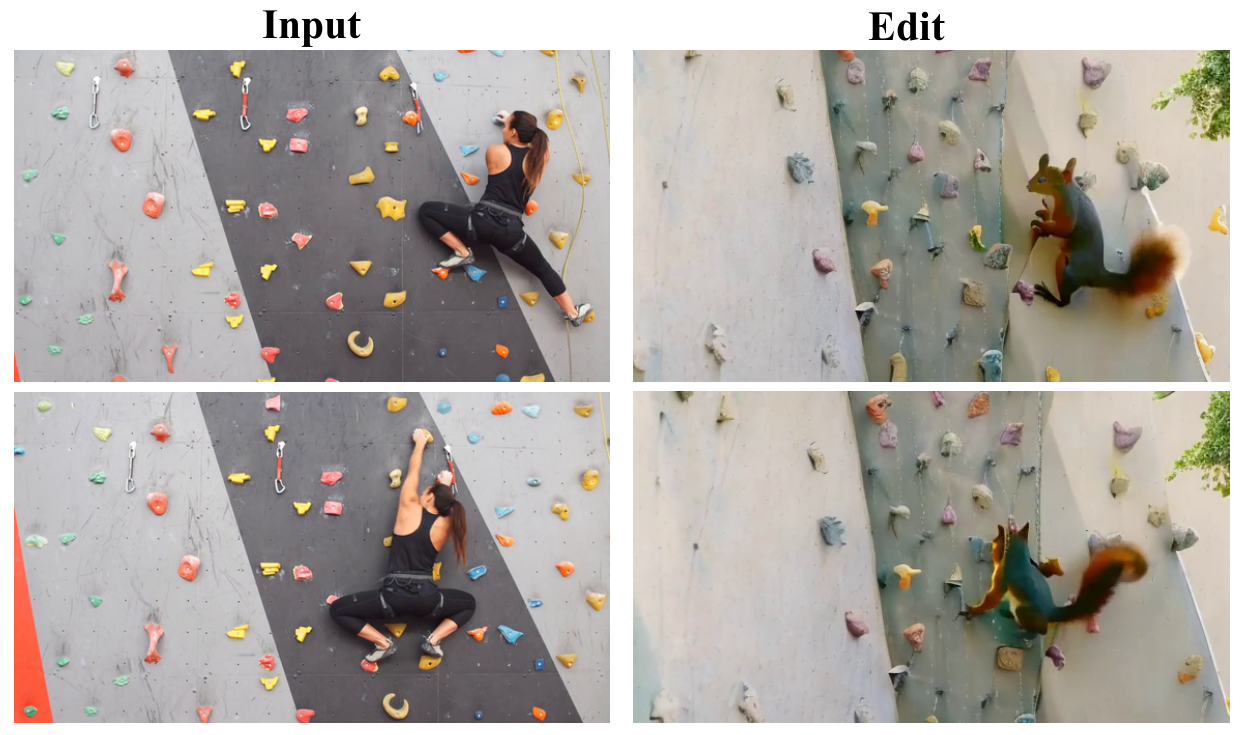}\vspace{-0.3cm}
    \caption{{\bf Limitations.} Our method struggles to preserve the original motion since the combination of the original motion and the edit prompt may be out of distribution for the T2V model.} 
    \label{fig:limitations}
    % \vspace{0.5cm}
\end{figure}

\subsection{Ablations}

In Fig.~\ref{fig:ablation}, we ablate key design choices in our framework. First, we consider alternative loss functions by substituting the pairwise SMM differences in Eq.\ref{eq:diff_objective} with: (i) full space-time features $\bs{f}$ and (ii) SMM features (Eq.\ref{eq:smm}). Figure~\ref{fig:ablation} (b) shows that space-time features restrict both shape and appearance variations; optimizing for SMM features (c) increases flexibility yet is insufficient for matching the edit.

We next ablate our guided sampling and latent initialization strategy. Sampling from the initial latent without optimization (d) converges to unrelated objects' pose, while initializing the optimization from randomly sampled noise (e) fails to retain the original motion characteristics.

\begin{table}
  \centering
  \begin{tabular}{l c}
    \toprule
    Method &  Judgements in our favor (\%)\\
    \midrule
    TokenFlow & 72.57\% \\
    Control-A-Video & 84.50\% \\
    Tune-A-Video & 77.80\% \\
    \bottomrule \vspace{-.6cm}
  \end{tabular}
  \caption{\textbf{User Study.} We report the percentage of judgments in our favour w.r.t. each baseline.}\afterfigure
  \label{table:amt}
\end{table}
\section{Discussion and conclusions} 
We tackled the task of text-driven motion transfer, focusing on scenarios where the source and target objects differ in shape and fine-grained motion traits. We introduced a  zero-shot method that utilizes a pre-trained text-to-video diffusion model, through a simple optimization framework.   Our work is the first to analyze and reveal new insights about space-time T2V features, and the first to show how to harness their properties for text-driven motion transfer.  

As for limitations, our performance relies on the generative priors learned by the T2V model. Thus, in some cases, the combination of target object and input video motion may be out-of-distribution for the T2V model. In this case, the motion fidelity of our results would be degraded or suffer from visual artifacts (Fig.~\ref{fig:limitations}). 
Furthermore, publicly available T2V models are still in infancy, in terms of quality, resolution, video length, and the scale of their training data compared to the vast distribution of natural videos. 
Despite the limitations of publicly available text-to-video models, our method achieves a significant improvement over prior state-of-the-art methods, demonstrating the potential of leveraging the priors and space-time feature space learned by these models. 

\section{Acknowledgement}
This project received funding from the Israeli Science Foundation (grant 2303/20). We thank GEN-1 authors for their help in conducting comparisons.

% \newpage
{\small
\bibliographystyle{ieeenat_fullname}
\bibliography{egbib}
}
\newpage
\appendix

\section{Text-to-Video Model Architecture and \\
Feature Selection}
\myparagraph{Text-to-Video Model.} 
We use ZeroScope \cite{zeroscope} text-to-video model, which is claimed to be fine-tuned from a Modelscope model \cite{wang2023modelscope} on video clips of the length of 24 frames and 576x320 resolution. Our generated results are in the same resolution with a length of 24 frames. The model was inflated from the StableDiffusion model \cite{stable_diffusion} by introducing temporal layers within each building block of the UNet.

\myparagraph{Feature Selection.}
The decoder of the UNet in ZeroScope comprises four blocks, each with a different resolution. We performed our analysis on coarse features, extracted from the 2nd decoder block of the UNet. 
We noticed that different coarse features in this block performed similarly for our task. Specifically, we tested intermediate features extracted from the spatial/temporal convolution models, output tokens from the spatial/temporal attention models, as well as features taken directly after the Upsampling block (a.k.a semantic DIFT features\cite{tang2023emergent}).
We empirically found that features extracted after the Upsampling block produce more visually appealing edit results.

\section{Implementation Details}

\myparagraph{Feature Extraction.}
To obtain intermediate latents, we follow \cite{pnpDiffusion2023} and use DDIM inversion (applying DDIM sampling in reverse order)
with a classifier-free guidance scale of 1 and 1000 forward steps, using a video-specific inversion prompt. We use these intermediate latents for initialization and extracting diffusion features.

\myparagraph{Initialization and Sampling.}
In our experiments, we use 50 denoising steps using Restart Sampling \cite{restart} combined with DDIM sampling \cite{song2020_ddim}, with a classifier-free guidance scale of 10. 
To obtain the initial noise, we apply the downsampling/upsampling operation $LF_{\xi}$, described in Eq. 4 with a factor $\xi=4$.

\myparagraph{Optimization details.} 
We apply the optimization described in Sec.~4.2 for the initial 20 denoising steps. In most of our experiments, we are using the Adam optimizer \cite{adam} with a learning rate of 0.01 for 30 optimization steps, but in cases where the edit required a bigger deviation from the original structure, we used a linear learning rate decay from 0.005 to 0.002 for 10 optimization steps.

\myparagraph{Runtime.}
The runtime of our method mainly consists of two parts - DDIM inversion, which takes \mytextapprox 10 minutes, and sampling with optimization, which takes \mytextapprox 7 minutes for 10 optimization steps per denoising step and \mytextapprox 15 minutes for 30 optimization steps per denoising step, depending on the configuration.

\section{Baseline Comparison Details}
For comparing with Tune-A-Video \cite{wu2022tuneavideo}, TokenFlow \cite{tokenflow2023} and  Control-A-Video \cite{chen2023controlavideo} we used the official repositories. For visual comparison with Gen-1 \cite{gen1}, we used the publicly available web platform. Since this platform outputs videos of different lengths with some frames being duplicated, we excluded Gen-1 from numerical comparisons. Since SA-NLA \cite{lee2023shape} takes $~10$ hours to train, we compare to their provided videos and edit prompts qualitatively.

\end{document}